\documentclass[10pt,a4paper,twocolumn]{article}
\usepackage[a4paper,top=19mm,bottom=21mm,left=17mm,right=17mm,columnsep=7mm]{geometry}
\usepackage[T1]{fontenc}
\usepackage[utf8]{inputenc}
\usepackage{times}
\usepackage{microtype}
\usepackage{amsmath,amssymb,mathtools}
\usepackage{graphicx,booktabs,array,tabularx,makecell,multirow}
\usepackage[dvipsnames]{xcolor}
\usepackage{tikz}
\usetikzlibrary{arrows.meta,positioning,calc,fit,shapes.geometric,backgrounds,patterns}
\usepackage{pgfplots}
\pgfplotsset{compat=1.17}
\usepackage{enumitem}
\usepackage{caption,subcaption}
\usepackage[sort&compress,numbers]{natbib}
\usepackage{fancyhdr}
\usepackage{url}
\usepackage{hyperref}
\usepackage{dblfloatfix}
\usepackage{float}
\usepackage{placeins}
\usepackage{balance}
\definecolor{LOrange}{HTML}{DF6800}
\definecolor{LDark}{HTML}{242424}
\definecolor{LGray}{HTML}{6F7780}
\definecolor{LCream}{HTML}{FDF6EC}
\hypersetup{colorlinks=true,citecolor=black,linkcolor=black,urlcolor=black,pdftitle={LarkupVideo: Towards Efficient and Grounded Agentic Long-Form Video Question Answering},pdfauthor={Abdelrahman Abounida and Sergey Yurchenko}}
\setlist{topsep=2pt,itemsep=1pt,parsep=0pt,leftmargin=1.2em}
\newcommand{\larkup}{LarkupVideo}

\newcommand{\LarkupTitle}{%
    \par\noindent
    \rule{\textwidth}{1.1pt}\par
    \vspace{7pt}
    \begin{center}
        {\Large\bfseries
        LarkupVideo: Towards Efficient and Grounded\\[2pt]
        Agentic Long-Form Video Question Answering\par}
    \end{center}
    \vspace{5pt}
    \noindent
    \rule{\textwidth}{1.1pt}\par
    \vspace{17pt}
    \begin{center}
        \begin{tabular}{@{}c@{\hspace{18mm}}c@{}}
            {\normalsize\bfseries Abdelrahman Abounida} &
            {\normalsize\bfseries Sergey Yurchenko} \\
            \small abdelrahmanaboneda@gmail.com &
            \small sergey-yur@outlook.com
        \end{tabular}
    \end{center}
    \vspace{16pt}
}

\renewenvironment{abstract}{%
    \par\begingroup
    \begin{center}
        \bfseries Abstract
    \end{center}
    \vspace{-3pt}
    \noindent\ignorespaces
}{%
    \par\endgroup
    \vspace{8pt}
}

\hypersetup{
    citecolor=blue,
    linkcolor=blue,
    urlcolor=blue
}

\begin{document}
\raggedbottom
\twocolumn[{\LarkupTitle}]
\suppressfloats[t]

\begin{abstract}

Long-form video question answering remains challenging
because relevant information can be scattered across
hours of footage, making exhaustive video processing
expensive and important details easy to miss. Existing
approaches struggle to balance computational efficiency
with accurate evidence retrieval, especially as video
duration increases and questions require reasoning
across different temporal scales
\cite{wu2024longvideobench,park2026toomanyframes}.

We introduce \textbf{\larkup{}}, an agentic multimodal
retrieval-augmented system that extends existing
vision-language models with persistent video memory
and query-aware reasoning. Rather than repeatedly
processing an entire video, \larkup{} builds a reusable,
timestamped index of speech and visual observations,
retrieves only the evidence relevant to each question,
and selectively re-inspects source segments when
additional evidence is needed. This enables targeted
retrieval across long recordings without requiring
the entire video to fit within a model's context window.

We evaluate \larkup{} in a curated pilot study of
predominantly long videos selected from Video-MME and
LongVideoBench and supplemented with additional
recordings, spanning diverse content types and temporal
structures. Multiple-choice answers are matched against
reference answers and open-ended answers are scored by an
LLM judge; question scores are weighted by difficulty and
aggregated into a video-level pass rate. Qwen 3.8 Max,
Gemini 3.8 Flash, Muse Spark 1.3, and MiniMax M3 are
evaluated both as \larkup{} vision readers and as
native baselines that process the full video, with TwelveLabs Pegasus 1.5
as an external baseline. \larkup{} achieves a higher
video-level pass rate with all four readers. Gemini
improves from 31.6\% to 78.9\% while reducing total
computational cost by 89.3\%, and Muse improves from
78.9\% to 84.2\% with a 35.8\% cost reduction, whereas
Qwen and MiniMax incur higher total costs. These
preliminary results suggest that
persistent indexing and agentic retrieval can make
long-form video understanding more accurate and, for
some models, more computationally efficient.

\end{abstract}

\section{Introduction}
\label{sec:intro}

Understanding long-form videos requires identifying
relevant information across extended periods of footage.
An answer to a question about an hour-long lecture,
documentary, or recording may depend on a brief visual
event, a spoken statement, or evidence scattered across
several distant segments. Although modern vision-language
models (VLMs) have demonstrated strong video understanding
capabilities, processing long recordings remains
computationally expensive. Analyzing the entire video
for every question introduces redundant computation,
while aggressive frame selection and summarization
risk discarding critical information. This creates
a fundamental trade-off between computational efficiency
and accurate evidence retrieval, particularly as videos
grow longer and questions require reasoning across
different temporal scales
\citep{wu2024longvideobench,fu2025videomme}.

Prior work addresses this trade-off through sparse
memory \citep{song2024moviechat}, hierarchical video
representations \citep{wang2025videotree}, and
retrieval-augmented or agent-based reasoning
\citep{luo2025videorag,fan2024videoagent}
(Section~\ref{sec:related}). An important question
nevertheless remains: can a system reuse previously
extracted video knowledge across multiple questions
while retaining the ability to recover information
that was missed or insufficiently observed? A fixed
representation can omit brief events, whereas repeated
source inspection can undermine the efficiency gained
through retrieval.

We address this question with \textbf{\larkup{}}, an
agentic multimodal retrieval-augmented system that
extends existing vision-language models with persistent
video memory. \larkup{} builds a reusable, timestamped
index of speech, visual observations, detected text, and
semantic embeddings once per video. For each question, an
investigation agent retrieves relevant evidence from this
index and, when the evidence is missing or ambiguous,
directs the underlying vision-language reader to
re-inspect selected intervals of the original recording.

Recovering missing information, however, does not
guarantee correct answers: evidence may be omitted during
indexing, missed during retrieval, misinterpreted during
re-inspection, or misused during answer synthesis.
Building on work that evaluates supporting evidence
beyond final-answer accuracy
\citep{huang2026egvqa,benami2026her,an2026agentvid},
we therefore also propose a four-stage diagnostic
framework for locating where evidence is lost.

We evaluate \larkup{} in a curated pilot study of
predominantly long videos from Video-MME, LongVideoBench,
and additional sources, using four vision-language models
both as \larkup{} readers and as native baselines that
answer directly from the full video. \larkup{} achieves a higher video-level pass
rate with all four readers, while its effect on total
computational cost depends on the reader
(Section~\ref{sec:results}).

Our work makes three main contributions:

\begin{enumerate}
    \item We introduce a persistent multimodal video
    indexing and agentic retrieval architecture that
    enables query-adaptive evidence selection and
    targeted source re-inspection
    (Section~\ref{sec:system}).

    \item We present a multi-model pilot evaluation of
    video-level pass rate and total computational cost,
    including an analysis of recurring failure cases
    (Sections~\ref{sec:setup}--\ref{sec:results}).

    \item We propose a stage-wise diagnostic framework
    for locating evidence loss across indexing,
    retrieval, re-inspection, and answer synthesis,
    as a basis for future controlled studies
    (Section~\ref{sec:diagnostics}).
\end{enumerate}

\section{Related Work}
\label{sec:related}

Long-form video understanding has increasingly shifted
from processing dense sequences of video frames toward
more compact representations of visual information.
LLoVi \citep{zhang2024llovi} demonstrates that short-clip
captions can be aggregated by language models to answer
questions about extended videos. MovieChat
\citep{song2024moviechat} introduces sparse memory
mechanisms for preserving information across longer
temporal contexts, while VideoTree
\citep{wang2025videotree} constructs hierarchical,
query-adaptive representations that progressively select
relevant visual information. These approaches reduce
the amount of video information presented to a reasoning
model. However, compression and frame selection introduce
an important trade-off: information excluded from the
representation may include the very evidence needed to
answer a subsequent question.

Retrieval-augmented and agent-based methods offer an
alternative by selecting evidence according to the
question being asked. Video-RAG \citep{luo2025videorag}
incorporates auxiliary information from speech,
recognized text, and object detection to improve
long-video comprehension. VideoAgent
\citep{fan2024videoagent} combines structured memory
with interactive tools for locating video segments
and querying object states. More recently, CARVE and
V-RAGBench \citep{lee2026vrag} demonstrate the importance
of selecting appropriate retrieval modalities and
temporal granularities while evaluating retrieval
separately from answer generation. These results
establish that video retrieval, structured memory,
and adaptive evidence selection are valuable techniques.
They also motivate a further question: how can a
reusable video representation support repeated queries
while allowing the system to recover evidence that
was missed during initial indexing or retrieval?

Recent benchmarks have made the importance of evidence
quality increasingly explicit. Video-MME
\citep{fu2025videomme} and LongVideoBench
\citep{wu2024longvideobench} evaluate video understanding
across diverse durations and reasoning tasks.
EG-VQA \citep{huang2026egvqa} extends evaluation
beyond answer accuracy by requiring temporally
grounded supporting evidence. HERBench
\citep{benami2026her} distinguishes failures to
retrieve relevant observations from failures to
integrate them, while AgentVidBench
\citep{an2026agentvid} introduces evidence milestones
for evaluating agent reasoning trajectories.
LongVidSearch \citep{yu2026longvidsearch} further
isolates multi-hop retrieval planning through a
standardized evidence-access interface. Together,
these studies show that answer correctness alone
does not fully characterize a video reasoning
system's ability to locate and use source evidence.

Efficiency introduces an additional challenge.
SAVE \citep{wilk2026save} evaluates agentic video reasoning under controlled model and computational budgets, showing that additional agent complexity does not consistently improve performance over simpler approaches or native inference. This motivates examining not only answer accuracy, but also the computational cost of acquiring supporting evidence.

Building on these directions, \larkup{} combines reusable multimodal indexing with an investigation agent that selects evidence and directs targeted visual re-inspection. The system separates investigation planning from visual observation, allowing different vision-language readers to operate within the same retrieval architecture. We further propose diagnosing evidence loss across indexing, retrieval, re-inspection, and answer synthesis. Our focus is on understanding how these components interact and how their contributions to accuracy and computational cost can be evaluated.

\section{System Architecture}
\label{sec:system}

The central design principle of \larkup{} is to separate video processing from question answering. A long recording may be queried many times, but each question can require different evidence. Processing the entire video for every question repeats much of the same computation, while relying only on a fixed summary risks losing details that become relevant later. We address this trade-off by constructing a reusable representation of the source and allowing an agent to return to the original video when additional evidence is needed.

\begin{figure*}[t]
\centering
\includegraphics[width=0.98\textwidth]{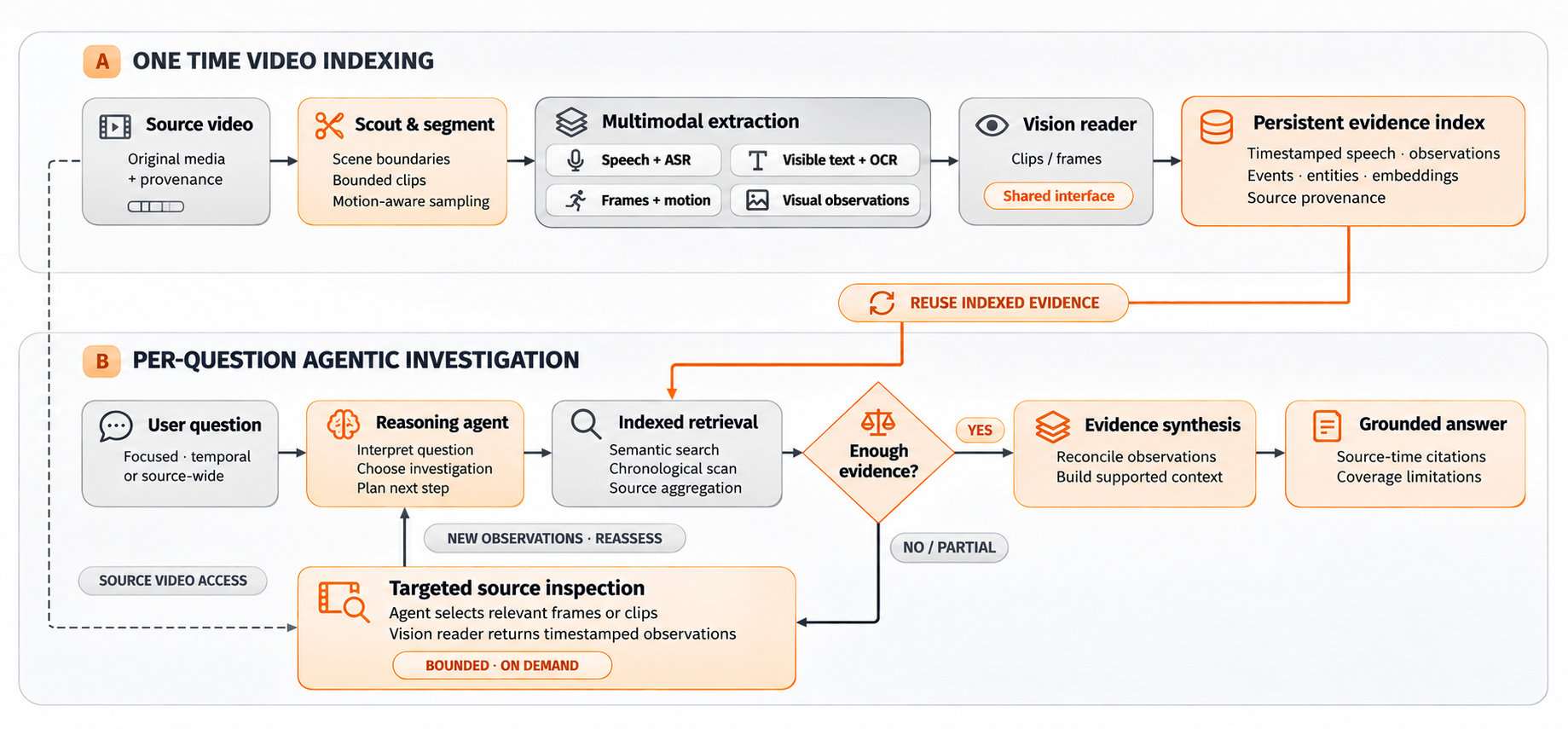}
\caption{\textbf{Overview of the \larkup{} architecture.}
A source video is processed once into a persistent, timestamped multimodal evidence index. For each question, an investigation agent selects a retrieval strategy, consults stored evidence, and can invoke targeted visual re-inspection of the original recording when additional information is required. Retrieved and newly observed evidence are combined to produce a grounded answer.}
\label{fig:arch}
\end{figure*}

As illustrated in Figure~\ref{fig:arch}, \larkup{} operates in two phases. First, a source video is processed into a persistent multimodal index containing timestamped speech, visual observations, detected text, and semantic representations. Second, an investigation agent interprets each question, retrieves relevant evidence, and determines whether additional visual inspection is necessary. This separates the cost of initially processing a recording from the work required to answer subsequent questions, without restricting the system to information captured during indexing.

\subsection{Constructing a Reusable Video Index}
\label{sec:indexing}

The indexing phase converts a video into smaller, source-aligned units that can be searched and reused. Processing begins by examining the media properties and identifying candidate scene boundaries. We combine cut-sensitive scene detection with fixed-duration windows so that recordings with few visible transitions do not produce excessively long analysis segments. This provides bounded units for visual inspection while retaining their positions in the original recording.

The system then extracts complementary forms of evidence. Speech transcription captures spoken information, while a vision-language reader describes selected clips and their visual content. The reader is accessed through a shared interface, so the same model can be used both during indexing and for targeted re-inspection at question time (Section~\ref{sec:reinspection}). Optical character recognition provides additional information from visible text, and representative frames preserve visual context associated with individual observations. These modalities are processed separately but linked through their timestamps, allowing the system to relate spoken statements, visible objects, and events occurring within the same interval.

A practical challenge is deciding which portions of a long recording to inspect during indexing. Uniform sampling provides temporal coverage but may allocate substantial computation to visually redundant material, whereas concentrating only on changing scenes can overlook information in relatively static footage. \larkup{} therefore uses motion-aware sampling that favors visual changes while keeping selections distributed across the source. The number of clips analyzed is bounded by an inspection budget determined by one of three processing modes:

\begin{itemize}
    \item \textbf{Fast:} samples sparsely and directs the agent to skip large portions of the source, trading coverage for speed.
    \item \textbf{Coverage:} lets the agent decide which segments to inspect or skip, balancing temporal coverage against cost.
    \item \textbf{Full:} processes the entire video densely, with minimal agent-driven selection.
\end{itemize}

\noindent All reported results use coverage mode (Section~\ref{sec:models}). In every mode, the index provides a structured account of the recording rather than an exhaustive representation of every frame.

The implementation supports configurable transcription and model-serving backends. Speech can be processed using faster-whisper or Deepgram, while multimodal embeddings can be generated using Qwen3-VL through supported serving adapters. The broader architecture does not depend on a single transcription service or vision-language model.

\subsection{Representing Persistent Video Evidence}

Extracted observations are organized into a persistent evidence store containing timestamped speech segments, visual descriptions, detected text, entities, summaries, and retrieval embeddings. Each record retains its source interval and evidence type, enabling subsequent questions to retrieve information without reconstructing the index.

We preserve the distinction between kinds of evidence because they do not support identical conclusions. A person mentioned in speech has not necessarily appeared on screen, observations of a person at two timestamps do not establish uninterrupted visibility between them, and two visually similar subjects cannot automatically be treated as the same individual. Source timestamps and record provenance allow later investigations to examine these distinctions rather than treating all indexed descriptions as interchangeable text.

Source-aligned records also support questions requiring evidence from multiple parts of a recording. A question about how an object changes over time, for example, may require observations from several distant intervals, which \larkup{} can retrieve and order chronologically.

Importantly, the evidence store persists across questions about the same video. New questions reuse previously extracted observations and embeddings, requesting additional inspection only when the stored information is insufficient. This reuse is the architectural basis for amortizing source-processing costs over multiple queries.

\subsection{Question-Conditioned Investigation}
\label{sec:investigation}

Rather than applying a fixed retrieval strategy to every
question, \larkup{} uses a reasoning agent to coordinate
evidence retrieval and visual inspection. In our
experiments, Claude Sonnet 5 serves as the investigation
agent, while a separate vision-language reader performs
visual observations. Given a question, the agent constructs
an investigation directive specifying the required evidence,
search scope, and operation. This separation allows the agent
to plan investigations independently of the selected vision
reader.

The investigation directive distinguishes three scopes:

\begin{enumerate}
    \item \textbf{Focused:}
    evidence associated with a particular
    subject, object, or event.

    \item \textbf{Temporal:}
    relationships between observations across
    time, including event sequences, transitions,
    and changes in state.

    \item \textbf{Source-wide:}
    evidence distributed throughout the
    recording, supporting comprehensive summaries
    and exhaustive inventories.
\end{enumerate}

The directive also specifies one of five investigation
goals:

\begin{enumerate}
    \item \textbf{Answer:}
    retrieve evidence to resolve a specific question.

    \item \textbf{Compare:}
    examine differences or similarities between
    subjects, events, or states.

    \item \textbf{Trace:}
    follow a subject, event, or state across
    multiple source intervals.

    \item \textbf{Enumerate:}
    identify all relevant occurrences
    or items within the requested scope.

    \item \textbf{Synthesize:}
    combine evidence from multiple intervals into
    a coherent explanation or summary.
\end{enumerate}

Additional constraints may specify relevant time
intervals, evidence modalities, record types, or
candidate answers supplied with the question.
Together, these parameters determine how evidence is
retrieved and examined based on the information
requested rather than the genre of the video.

The choice of retrieval strategy matters because
relevance alone does not guarantee sufficient coverage.
For a question about a particular visible object,
ranked retrieval may identify the most informative
observations. In contrast, a question asking for every
item appearing in source-authored material requires a
complete traversal of the corresponding records.
\larkup{} therefore supports chronological and
paginated retrieval alongside similarity-based search.
Temporal questions likewise require attention to event
order, since an initial occurrence, a repeated occurrence,
and a later reference to an event may produce similar
descriptions while supporting different answers.

\subsection{Re-Inspection and Answer Synthesis}
\label{sec:reinspection}

The persistent index provides the initial evidence for answering a question, but it cannot be assumed to contain every relevant visual detail. A short event may fall between sampled frames, a visual description may omit an attribute, or several observations may leave the requested sequence ambiguous. To address these cases, \larkup{} allows the investigation agent to inspect selected portions of the original recording during question answering.

After retrieving indexed evidence, the agent assesses whether the observations are sufficient. When information is missing, ambiguous, or contradictory, it selects relevant source intervals and specifies the visual details to inspect. The vision-language reader then analyzes the requested clips or frames and returns observations associated with source timestamps. For example, a question about an object's changing appearance may require observations before and after a transition, while identifying a recurring subject may require comparing distinguishing attributes across intervals. When the remaining budget allows, the agent can reassess the returned evidence and request further inspection.

Unlike a fixed video summary, the persistent index therefore retains a path to the original recording. Recovery nevertheless depends on selecting appropriate intervals and correctly interpreting their contents. The implementation supports local and managed inspection mechanisms, whose execution paths are not assumed to be equivalent.

New observations retain their source timestamps and inspection coverage and are combined with retrieved evidence during answer synthesis. Chronological observations and source-established attributes help distinguish identities and states across intervals, while temporal coverage limits claims of continuous presence or absence. If the available evidence cannot establish a claim, the system expresses uncertainty or qualifies its inspection coverage rather than treating missing observations as proof of non-occurrence.
\section{Evaluation Setup}
\label{sec:setup}

We evaluate \larkup{} on a small, curated collection of
long-form video question-answering tasks, measuring the
video-level pass rate and total computational cost of each
configuration. The study is an exploratory pilot rather
than a fully controlled architectural comparison: the
evaluation set was used during development, and execution
settings changed across runs (Section~\ref{sec:models}).
The resulting limits on interpretation are discussed in
Section~\ref{sec:limits-eval}.

\subsection{Evaluation Set}
\label{sec:dataset}

The evaluation questions were gathered from established
video-understanding benchmarks, namely Video-MME and
LongVideoBench \citep{fu2025videomme,wu2024longvideobench},
together with supplementary questions on additional
videos. The resulting set contains 19 videos spanning
instructional content, animated explainers, presentations,
news segments, and performances. Most recordings are long:
17 of the 19 videos last between 30 minutes and 3 hours.
This is a curated subset rather than either benchmark's
complete official test set, and it serves as a development
and diagnostic set rather than a held-out benchmark.

Each video is associated with several questions in
multiple-choice or open-ended format; the number and mix
of questions vary across videos. The questions examine
visual attributes, event ordering, scene recognition,
changes in subject or object state, and information
distributed across a recording. Some require identifying
a specific moment, while others involve comparing events,
following temporal changes, or synthesizing information
from multiple intervals.

\begin{table}[t]
\centering
\caption{Overview of the pilot evaluation.}
\label{tab:setup}
\small
\begin{tabularx}{\linewidth}{@{}lX@{}}
\toprule
Component & Setting \\
\midrule
Question sources & Video-MME; LongVideoBench; supplementary questions \\
Videos & 19 (17 between 30\,min and 3\,h) \\
Question formats & Multiple-choice and open-ended \\
Vision readers & Qwen, Muse, Gemini, MiniMax \\
Reasoning agent & Claude Sonnet 5 (\larkup{} only) \\
Processing mode & Coverage \\
LLM judge & DeepSeek V4 Pro \\
Difficulty weights & 1 / 3 / 5 (easy / medium / hard) \\
Video pass threshold & $\tau = 0.85$ \\
External baseline & TwelveLabs Pegasus 1.5 \\
\bottomrule
\end{tabularx}
\end{table}

\subsection{Models and Execution}
\label{sec:models}

We evaluate four vision-language models: Qwen 3.8 Max,
Muse Spark 1.3, Gemini 3.8 Flash, and MiniMax M3. Each is
evaluated in two configurations:

\begin{itemize}
    \item \textbf{Native baseline:} the full video is
    sent directly to the vision model through its API, and
    the model answers each question on its own. No
    reasoning agent, persistent index, or targeted
    re-inspection is involved.
    \item \textbf{\larkup{}:} the vision model serves as
    the reader within the full pipeline of
    Section~\ref{sec:system}, run in coverage mode, with
    Claude Sonnet 5 as the investigation agent that selects
    which segments and frames the reader analyzes.
\end{itemize}

\noindent The comparison therefore measures the
\larkup{} pipeline as a whole, including the addition of
a reasoning agent, against direct full-video inference;
it does not isolate the contribution of indexing and
retrieval from that of the agent. We additionally include TwelveLabs Pegasus 1.5 as an
external video-understanding baseline, evaluated only
through its native interface.

Native workflows differ in the video length, audio, and
frames they can accept. For Gemini 3.8 Flash, a compact
model, the API accepted the full recordings, but native
answers frequently failed on very long videos of roughly
two hours or more. Within \larkup{}, the reader analyzes
only bounded segments selected from the persistent index
rather than an entire long recording at once.
We return to the implications of this difference in
Section~\ref{sec:results-score}.

Native and \larkup{} runs share question identifiers,
enabling paired comparison of outcomes. However, the
\larkup{} records span development iterations with changes
to the index, timeout settings, and inspection strategy,
and include retries.

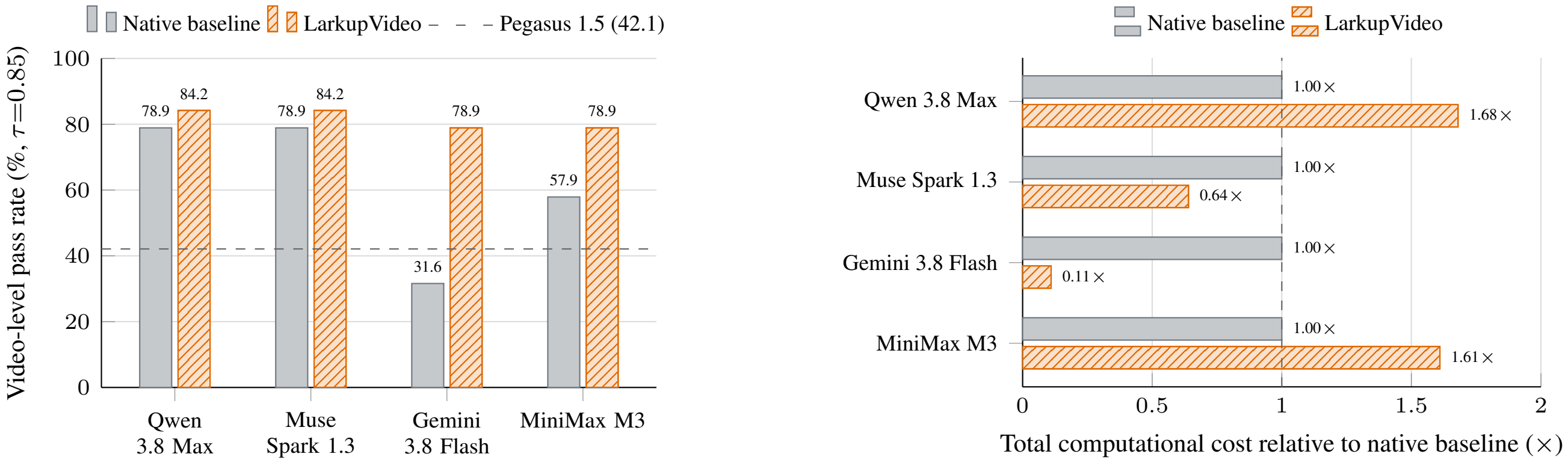
\begin{figure*}[t]
\centering
\begin{minipage}[b]{0.45\textwidth}
\centering
\begin{tikzpicture}
\begin{axis}[
    width=0.97\linewidth, height=5.2cm,
    ybar, bar width=10pt,
    ymin=0, ymax=100,
    ylabel={Video-level pass rate (\%, $\tau{=}0.85$)},
    ylabel style={font=\footnotesize},
    yticklabel style={font=\scriptsize},
    symbolic x coords={Qwen,Muse,Gemini,MiniMax},
    xtick={Qwen,Muse,Gemini,MiniMax},
    xticklabels={Qwen 3.8 Max, Muse Spark 1.3, Gemini 3.8 Flash, MiniMax M3},
    xticklabel style={font=\scriptsize, align=center, text width=1.5cm},
    enlarge x limits=0.18,
    point meta=explicit symbolic,
    nodes near coords,
    nodes near coords style={font=\tiny},
    legend style={font=\scriptsize, at={(0.5,1.03)}, anchor=south, legend columns=-1, draw=none},
    ymajorgrids, grid style={LGray!25},
    axis lines*=left,
]
\addplot[fill=LGray!40, draw=LGray]
    coordinates {(Qwen,78.9) [78.9] (Muse,78.9) [78.9] (Gemini,31.6) [31.6] (MiniMax,57.9) [57.9]};
\addplot[fill=LOrange!20, draw=LOrange, postaction={pattern=north east lines, pattern color=LOrange}]
    coordinates {(Qwen,84.2) [84.2] (Muse,84.2) [84.2] (Gemini,78.9) [78.9] (MiniMax,78.9) [78.9]};
\draw[dashed, LDark!70] ({rel axis cs:0,0}|-{axis cs:Qwen,42.1}) -- ({rel axis cs:1,0}|-{axis cs:Qwen,42.1});
\addlegendimage{dashed, LDark!70, line legend}
\legend{Native baseline, \larkup{}, Pegasus 1.5 (42.1)}
\end{axis}
\end{tikzpicture}
\end{minipage}\hfill
\begin{minipage}[b]{0.47\textwidth}
\centering
\begin{tikzpicture}
\begin{axis}[
    width=0.88\linewidth, height=5.2cm,
    xbar, bar width=7pt,
    xmin=0, xmax=2,
    xlabel={Total computational cost relative to native baseline ($\times$)},
    xlabel style={font=\footnotesize},
    xticklabel style={font=\scriptsize},
    symbolic y coords={MiniMax,Gemini,Muse,Qwen},
    ytick={MiniMax,Gemini,Muse,Qwen},
    yticklabels={MiniMax M3, Gemini 3.8 Flash, Muse Spark 1.3, Qwen 3.8 Max},
    yticklabel style={font=\scriptsize},
    enlarge y limits=0.18,
    point meta=explicit symbolic,
    nodes near coords,
    nodes near coords align={horizontal},
    every node near coord/.append style={anchor=west},
    nodes near coords style={font=\tiny},
    legend style={font=\scriptsize, at={(0.5,1.03)}, anchor=south, legend columns=-1, draw=none},
    xmajorgrids, grid style={LGray!25},
    axis lines*=left,
    reverse legend,
]
\addplot[fill=LOrange!20, draw=LOrange, postaction={pattern=north east lines, pattern color=LOrange}]
    coordinates {(1.68,Qwen) [1.68$\times$] (0.64,Muse) [0.64$\times$] (0.11,Gemini) [0.11$\times$] (1.61,MiniMax) [1.61$\times$]};
\addplot[fill=LGray!40, draw=LGray]
    coordinates {(1.00,Qwen) [1.00$\times$] (1.00,Muse) [1.00$\times$] (1.00,Gemini) [1.00$\times$] (1.00,MiniMax) [1.00$\times$]};
\draw[dashed, LDark!70] ({axis cs:1,Qwen}|-{rel axis cs:0,0}) -- ({axis cs:1,Qwen}|-{rel axis cs:0,1});
\legend{\larkup{}, Native baseline}
\end{axis}
\end{tikzpicture}
\end{minipage}
\caption{
\textbf{Video-level pass rate and relative total
computational cost.}
Left: percentage of the 19 videos whose difficulty-weighted
score reaches $\tau = 0.85$, for the native baseline and
\larkup{}; the dashed line marks the native-only Pegasus 1.5
baseline. Right: total computational cost of each \larkup{}
configuration relative to the native baseline of the same
reader ($1.00\times$, dashed line). Hatched orange bars
denote \larkup{}.
}
\label{fig:pilotcomparison}
\end{figure*}

\subsection{Scoring}
\label{sec:scoring}

\paragraph{Question correctness.}
Each question $i$ of video $v$ receives a binary
correctness score $s_{v,i} \in \{0,1\}$. Multiple-choice
predictions are compared against their reference answers.
Open-ended responses are assessed against the expected
reference answers by DeepSeek V4 Pro acting as an
automated LLM judge. The judge compares each
response with the reference answer. When this comparison
yields only a partial match, the judge performs a
follow-up evaluation, repeated at most twice, to determine
whether the response adequately addresses the question. This follow-up is part
of scoring, not of question answering: it re-examines the
existing response against the question and reference
answer, without querying the evaluated system again or
changing its answer. The verdict reached after the
follow-up is used as the final correctness score. All
open-ended correctness judgments rely on the LLM judge and
were not verified by human annotators, while
multiple-choice answers are evaluated directly against
their references.

\paragraph{Difficulty weighting.}
The LLM judge also classifies each question as easy,
medium, or hard, with weights $w_{v,i} \in \{1, 3, 5\}$,
respectively. These weights are a heuristic choice that
gives greater importance to questions requiring deeper
reasoning or more detailed analysis; they were not
statistically optimized or experimentally validated. The
difficulty-weighted score of video $v$ is
\begin{equation}
S_v = \frac{\sum_{i=1}^{n_v} w_{v,i}\, s_{v,i}}
           {\sum_{i=1}^{n_v} w_{v,i}},
\label{eq:videoscore}
\end{equation}
where $n_v$ is the number of questions for video $v$.
Because hard questions carry more weight, an incorrect
answer to a hard question lowers $S_v$ more than an
incorrect answer to an easy one.

\paragraph{Video-level pass rate.}
A video passes when $S_v \geq \tau$, with $\tau = 0.85$.
This threshold was also selected heuristically: it
requires a high level of correctness while allowing
limited tolerance for errors on individual questions.
For example, for a video with one easy, one medium, and
one hard question, missing only the easy question gives
$S_v = 8/9 \approx 0.89$ and the video passes, whereas
missing only the hard question gives
$S_v = 4/9 \approx 0.44$ and the video fails. The
video-level pass rate is the percentage of videos that
pass:
\begin{equation}
\mathrm{Pass\ rate} = \frac{100\%}{V}
\sum_{v=1}^{V} \mathbb{1}\!\left[S_v \geq \tau\right],
\label{eq:score}
\end{equation}
where $V = 19$. Each video contributes equally,
regardless of its number of questions, so pass rates are
multiples of $1/19$ (about 5.3 percentage points), and
both question formats contribute to the same score.

\subsection{Cost Measurement}
\label{sec:cost}

We report total computational cost as the recorded API
expenditure of a complete benchmark run for each
configuration:

\begin{itemize}
    \item \textbf{Native baseline:} API usage associated
    with submitting the video content and obtaining the
    generated answer, including applicable additional
    requests.
    \item \textbf{\larkup{}:} all expenditure of the
    pipeline, comprising initial video indexing, audio
    transcription, vision-reader calls during indexing and
    re-inspection, investigation-agent reasoning,
    retrieval, question answering, and associated
    execution costs. This total therefore
    combines the cost of building the video representation
    with the cost of using it to answer questions.
    \item \textbf{Pegasus 1.5:} native video-processing
    and answer-generation API usage.
\end{itemize}

\noindent The reported totals cover system execution only;
the cost of LLM-judge evaluation is excluded for all
configurations. System follow-up requests and retries are included
when they appear in the recorded billing data, and cached
operations are counted only when they were billed. Missing
cost records are not treated as zero expenditure. Because
the available records combine provider-reported charges
and exported estimates of varying completeness, we report
each \larkup{} configuration's total cost relative to the
native baseline of the same reader rather than currency
amounts. Pegasus 1.5 has no corresponding \larkup{}
configuration, so no relative cost is reported for it.
\section{Results}
\label{sec:results}

Table~\ref{tab:pilotresults} and
Figure~\ref{fig:pilotcomparison} report the video-level
pass rate and relative total computational cost of the
four vision-language models in their native-baseline and
\larkup{} configurations. \larkup{} achieves a higher pass
rate with all four readers, while its effect on total cost
differs across models.

\subsection{Video-Level Pass Rate}
\label{sec:results-score}

Qwen 3.8 Max and Muse Spark 1.3 each improve from 78.9\%
to 84.2\%, a gain of 5.3 percentage points that
corresponds to one additional video passing. MiniMax M3
improves from 57.9\% to 78.9\% (+21.0 points), and the
external Pegasus 1.5 baseline reaches 42.1\%.

\begin{table}[!htb]
\centering
\caption{Video-level pass rate ($\tau = 0.85$) and change
in total computational cost relative to the native
baseline; negative values indicate cost reductions.
Pegasus 1.5 is evaluated natively only.}
\label{tab:pilotresults}
\small
\setlength{\tabcolsep}{3pt}
\begin{tabularx}{\linewidth}{@{}Xrrr@{}}
\toprule
Reader & Native & \larkup{} & Cost change \\
\midrule
Qwen 3.8 Max & 78.9\% & 84.2\% & $+$67.6\% \\
Muse Spark 1.3 & 78.9\% & 84.2\% & $-$35.8\% \\
Gemini 3.8 Flash & 31.6\% & 78.9\% & $-$89.3\% \\
MiniMax M3 & 57.9\% & 78.9\% & $+$61.1\% \\
\midrule
Pegasus 1.5 & 42.1\% & -- & -- \\
\bottomrule
\end{tabularx}
\end{table}

The largest gain is observed for Gemini 3.8 Flash, which
improves from 31.6\% to 78.9\% (+47.3 points). Its low
native pass rate coincides with frequent failures on very
long recordings of roughly two hours or more
(Section~\ref{sec:models}); the records show this drop in
performance but do not identify its internal cause. Within \larkup{},
Gemini analyzes shorter, targeted segments selected from
the persistent index rather than processing each full
video for every question. The observed improvement therefore
reflects the combined effect of long-video accessibility,
persistent indexing, and question-conditioned retrieval,
and should not be attributed to improved retrieval or
reasoning alone. The available records do not establish
that every native failure resulted from video length.

Because the pass rate combines multiple-choice and
open-ended questions and applies a video-level threshold,
it is not directly comparable to per-question accuracy,
and the results do not show whether the gains are
concentrated in one question format.

\subsection{Total Computational Cost}
\label{sec:results-cost}

Total cost follows a different pattern from the pass
rate. Gemini reduces total cost by 89.3\% while gaining
47.3 points, and Muse reduces it by 35.8\% while gaining
5.3 points; both thus improve the performance--cost
trade-off relative to their native baselines. In
contrast, Qwen requires 67.6\% more expenditure for a
5.3-point gain, and MiniMax requires 61.1\% more for a
21.0-point gain.

The reductions are consistent with the intended advantage
of building a video representation once and reusing it
instead of repeatedly processing long source inputs.
However, the \larkup{} totals combine indexing,
transcription, retrieval, agent, re-inspection, and retry
expenditure that were not separated under matched
conditions, so the differences cannot be attributed to
index reuse alone. The computational efficiency of
\larkup{} therefore depends on the underlying vision
reader.

\subsection{Recurring Failure Cases}
\label{sec:failures}

Failures overlap substantially across configurations:
a small subset of particularly challenging videos fails
to pass in most native and \larkup{} configurations. These failures are associated mainly
with questions that require deep visual analysis: small
objects, subtle attributes, brief events, or highly
specific details within individual frames. For such
questions, persistent indexing and targeted retrieval can
locate the relevant interval and make it accessible to
the reader, but they do not guarantee that the
vision-language reader interprets fine-grained visual
evidence correctly. Improving accessibility and improving
visual interpretation are therefore distinct problems; we
discuss directions for the latter in
Section~\ref{sec:future}.
\section{Proposed Evidence-Loss Diagnostics}
\label{sec:diagnostics}

The video-level pass rate shows whether \larkup{} answers
correctly, but not why it fails. An incorrect prediction
may stem from missing information during indexing,
unsuccessful retrieval, incomplete visual inspection, or
misuse of available evidence, and each requires a
different remedy. We therefore propose a stage-wise
diagnostic protocol based on independently annotated
source intervals and recorded execution traces. The
current pilot lacks the complete annotations needed to
apply it, so we present it as methodology for future
evaluation rather than as measured results.

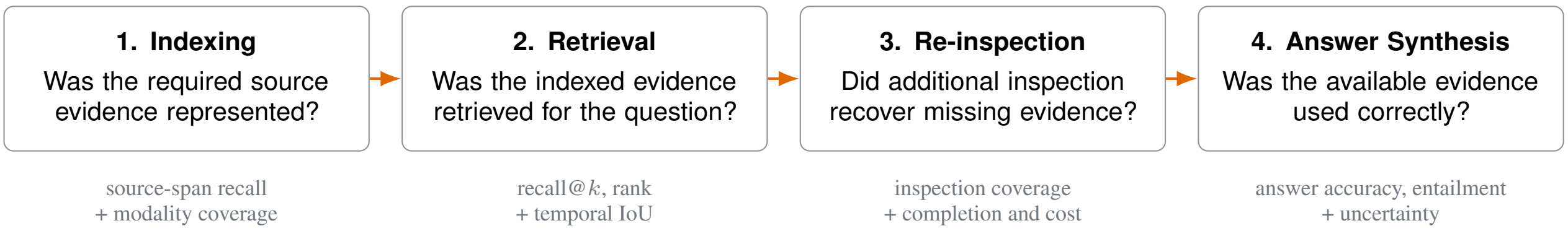
\begin{figure*}[t]
\centering
\resizebox{.95\textwidth}{!}{\begin{tikzpicture}[
    >=Latex,
    font=\sffamily\footnotesize,
    box/.style={
        draw=LGray!80,
        fill=white,
        rounded corners=3pt,
        text width=3.8cm,
        minimum height=1.6cm,
        align=center
    },
    arr/.style={
        -Latex,
        thick,
        draw=LOrange
    },
    metric/.style={
        font=\scriptsize,
        color=LGray,
        align=center
    }
]

\node[box] (i) at (0,0) {
    \textbf{1. Indexing}\\[3pt]
    Was the required source\\
    evidence represented?
};

\node[box] (r) at (4.4,0) {
    \textbf{2. Retrieval}\\[3pt]
    Was the indexed evidence\\
    retrieved for the question?
};

\node[box] (v) at (8.8,0) {
    \textbf{3. Re-inspection}\\[3pt]
    Did additional inspection\\
    recover missing evidence?
};

\node[box] (a) at (13.2,0) {
    \textbf{4. Answer Synthesis}\\[3pt]
    Was the available evidence\\
    used correctly?
};

\draw[arr] (i.east) -- (r.west);
\draw[arr] (r.east) -- (v.west);
\draw[arr] (v.east) -- (a.west);

\node[metric] at (0,-1.37) {
    source-span recall\\
    + modality coverage
};

\node[metric] at (4.4,-1.37) {
    recall@$k$, rank\\
    + temporal IoU
};

\node[metric] at (8.8,-1.37) {
    inspection coverage\\
    + completion and cost
};

\node[metric] at (13.2,-1.37) {
    answer accuracy, entailment\\
    + uncertainty
};

\end{tikzpicture}}
\caption{\textbf{Stage-wise evidence-loss diagnostics.}
The proposed framework identifies failures in indexing, retrieval, re-inspection, and answer synthesis using independently annotated source evidence and recorded execution traces. The diagram shows methodology, not measured results.}
\label{fig:stages}
\end{figure*}

\subsection{Grounding Questions in Source Evidence}

For a question $q$ about video $v$, let
\[
G_q = \{(s_j,e_j,m_j,z_j)\}
\]
be the set of independently annotated supporting
observations, where $[s_j,e_j]$ is a source interval,
$m_j$ an evidence modality (speech, visible text, video,
or a combination), and $z_j$ the required event or
subject-state relation. Some questions need a single
observation; others, such as a change in an object's
appearance, need several, and recovering only one does
not establish the answer. Annotations should therefore
identify all necessary evidence and flag ambiguous or
reference-conflicted questions before system predictions
are inspected.

Let $I_v$ denote the persistent index, $R_q(B)$ the
evidence retrieved under budget $B$, $W_q$ the completed
re-inspection observations, and $A_q$ the final answer.
These records determine whether each necessary
observation was available at each stage. Matching is
based on source-supported facts and temporal
relationships rather than textual similarity, since
different descriptions may refer to the same event.

\subsection{Failure Stages}

Following the architecture in Section~\ref{sec:system},
the protocol distinguishes four failure stages:

\begin{enumerate}
    \item \textbf{Indexing failure:} observable information is missing from the index, for example because sampling omitted a short event or a description dropped a detail.

    \item \textbf{Retrieval failure:} the evidence exists in the index but is not returned within the retrieval budget, due to poor semantic matching, an inappropriate temporal scope, or incomplete coverage.

    \item \textbf{Re-inspection failure:} evidence is insufficient, but inspection is not triggered, targets the wrong intervals, or fails to resolve the gap. Execution failures would be recorded separately from incorrect visual interpretation.

    \item \textbf{Answer synthesis failure:} sufficient evidence reaches the answer module, but the prediction is incorrect, unsupported, or loses temporal or subject distinctions.
\end{enumerate}

These categories are not exclusive, so the protocol would
record the first identifiable failure stage together with
any subsequent ones, tracking output-format violations
separately. To
confirm a bottleneck, annotated evidence can be injected
after indexing, retrieval, or re-inspection while all
other conditions are held fixed. If an error disappears,
this provides stronger evidence about its cause than the
final prediction alone.

\subsection{Metrics and Ablations}

Retrieval can be measured with recall@$k$, precision@$k$,
and mean reciprocal rank, and temporal localization with
temporal intersection-over-union (tIoU). For multi-evidence
questions, the protocol would additionally measure whether the complete
set of necessary evidence was recovered. Answer-level
evaluation considers correctness, consistency with
supporting evidence, and appropriate uncertainty under
incomplete coverage; failing to retrieve an observation
must not be treated as proof of absence. Execution
measurements include attempted and completed inspection
calls, processing time, token usage, and cost.

To isolate component contributions, we propose four
configurations evaluated under matched models, prompts,
indexes, scoring rules, and budgets:

\begin{enumerate}
    \renewcommand{\labelenumi}{\Alph{enumi}.}
    \item \textbf{Index-only:} retrieval from the persistent index with a fixed answer model.
    \item \textbf{Agentic retrieval:} question-conditioned retrieval without visual re-inspection.
    \item \textbf{Full \larkup{}:} agentic retrieval with bounded re-inspection.
    \item \textbf{Native baseline:} the same vision model answering directly from the full video, as in Section~\ref{sec:models}.
\end{enumerate}

\noindent Systems with different backbones, such as
Pegasus 1.5, remain external references rather than
controlled comparisons. Cost comparisons should include
both per-question execution and reusable indexing costs,
and statistical tests should preserve question-level
pairing and account for clustering by video.
\section{Discussion and Future Work}
\label{sec:discussion}

The pilot results suggest that separating video processing
from question answering can make long-form video question
answering both more reliable and, for some readers, less
expensive. The variation in total cost across readers
shows, however, that this benefit is not automatic: the
efficiency of \larkup{} depends on the initial indexing
cost, the number of questions asked about each video, and
the amount of re-inspection required. The current
evaluation does not establish a break-even point between
indexing cost and repeated queries.

\subsection{Current Limitations}
\label{sec:limits-system}

The recurring failure cases (Section~\ref{sec:failures})
point to the main system-level limitation: fine-grained
visual evidence. Even when the agent identifies the
relevant temporal region, re-inspection may miss the exact
frames containing a brief action or subtle transition,
and the reader may lack the resolution or capability to
interpret small objects and detailed attributes. These
difficulties grow with video duration unless the
inspection budget grows accordingly. Observations at
separate timestamps also cannot establish uninterrupted
presence or prove that an event never occurred.

Development additionally revealed execution issues,
including incomplete inspection coverage, unavailable
visual observations, and request timeouts. These can
prevent relevant evidence from reaching the agent even
when the correct interval has been identified.

\subsection{Evaluation Limitations}
\label{sec:limits-eval}

The pilot evaluation supports descriptive rather than
strictly causal conclusions. Because the native baseline
uses no reasoning agent, the gains reflect the full
\larkup{} pipeline, including the agent, rather than
indexing and retrieval alone. Differences in execution
settings, index versions, retries, and model input
capabilities across runs further prevent attributing
improvements to individual components, and the 19-video set is small
and was used during development.

The scoring procedure introduces further limitations.
Both question difficulty and the correctness of
open-ended answers were determined by the same LLM judge
without human validation, so misclassified difficulty or
incorrect verdicts propagate directly into the pass rate.
Follow-up judge evaluations are triggered only by partial
matches and are performed by the same judge, so borderline
verdicts depend on a single automated evaluator. Finally, the difficulty
weights and the pass threshold are heuristic choices, and
results may differ under other settings.

\subsection{Future Directions}
\label{sec:future}

\paragraph{Fine-grained visual inspection.}
Agent-guided inspection could become adaptive and
coarse-to-fine: the agent would first identify broad
intervals, progressively narrow the temporal search, and
request denser sampling or higher-resolution frames when
small details matter. Together with improved temporal
localization and adaptive frame selection, this directly
targets the recurring failure cases. Caching previously
examined intervals could further reduce model calls and
latency.

\paragraph{Indexing efficiency.}
Some indexed frames remain redundant despite scene
detection and motion-aware sampling. Selecting
observations by visual novelty, scene structure, speech
activity, and temporal diversity could reduce indexing
time, storage, and cost, provided that sparser sampling
does not miss short-lived events.

\paragraph{Controlled evaluation.}
We plan controlled ablations comparing index-only
retrieval, agentic retrieval without re-inspection, and
the full \larkup{} pipeline under matched models, prompts,
and budgets (Section~\ref{sec:diagnostics}), together with
a native baseline that uses the same reasoning agent, to
separate the agent's contribution from that of indexing
and retrieval. Future
analyses will also report paired statistical uncertainty
for the 19 videos, such as confidence intervals for
pass-rate differences; sensitivity to the pass threshold
(e.g., $\tau \in \{0.75, 0.85, 0.95\}$) and to alternative
difficulty weights; and evidence-grounding metrics based
on the proposed diagnostic framework. None of these
analyses is part of the present pilot.

\paragraph{Broader tasks and benchmarks.}
Beyond conventional question answering, we plan to study
tasks that demand exhaustive coverage, such as retrieving
every mention of a concept in a five-hour lecture,
extracting all slide text, or reconstructing the
chronological sequence of explanations. Evaluation should
also expand to larger benchmarks with multi-hour
recordings, multi-hop reasoning, and fine-grained visual
questions to test whether the pilot observations
generalize.

\subsection{Reproducibility}

Reproducibility depends on access to source videos, model
interfaces, and execution configurations. The accompanying
evaluation materials retain scoring records, source
identifiers, and available request metadata without
redistributing full videos or protected frames, in
accordance with the licensing conditions of the original
benchmarks and media. Future evaluations should preserve
model configurations and execution records to separate
reproducible findings from effects of changing hosted
services.

\section{Conclusion}
\label{sec:conclusion}

We presented \larkup{}, an agentic multimodal
retrieval-augmented system for long-form video question
answering that combines reusable video indexing,
question-conditioned retrieval, and targeted visual
re-inspection. In a pilot evaluation on 19 predominantly
long videos, \larkup{} achieved a higher video-level pass
rate than native full-video inference with all four
vision readers, while its effect on total computational
cost varied by reader. Recurring failures on fine-grained
visual questions show that making evidence accessible does
not by itself ensure that it is interpreted correctly. We
also proposed a stage-wise diagnostic framework for
locating evidence loss. These findings remain preliminary;
controlled ablations, statistical uncertainty analysis,
and larger benchmarks are needed to establish how
agent-directed evidence access can make long-form video
understanding more accurate, efficient, and reliably
grounded.
\bibliographystyle{unsrtnat}
\bibliography{references}
\appendix

\end{document}